\documentclass[nohyperref]{article}

\usepackage{microtype}
\usepackage[T1]{fontenc}
\usepackage{graphicx}
\usepackage{booktabs} %

\usepackage{hyperref}

\usepackage[accepted]{icml2022}

\usepackage{amsmath}
\usepackage{amssymb}
\usepackage{mathtools}
\usepackage{amsthm}

\usepackage[capitalize,noabbrev]{cleveref}

\theoremstyle{plain}

\theoremstyle{definition}

\theoremstyle{remark}

\usepackage[textsize=tiny]{todonotes}

\usepackage{tikz}

\usepackage{pgfplots} %
\usepackage{csquotes}
\usepackage{caption}
\usepackage{subcaption}

\icmltitlerunning{Reducing Exploitability with Population Based Training}

\begin{document}

\twocolumn[
\icmltitle{Reducing Exploitability with Population Based Training}

\icmlsetsymbol{equal}{*}

\begin{icmlauthorlist}
\icmlauthor{Pavel Czempin}{ucb,tum}
\icmlauthor{Adam Gleave}{ucb}
\end{icmlauthorlist}

\icmlaffiliation{ucb}{University of California, Berkeley}
\icmlaffiliation{tum}{Technical University of Munich}

\icmlcorrespondingauthor{Pavel Czempin}{pavel.czempin@berkeley.edu}

\icmlkeywords{Machine Learning, ICML, Reinforcement Learning, RL, PBT}

\vskip 0.3in
]

\printAffiliationsAndNotice{}  %

\begin{abstract}
Self-play reinforcement learning has achieved state-of-the-art, and often superhuman, performance in a variety of zero-sum games.
Yet prior work has found that policies that are highly capable against regular opponents can fail catastrophically against \emph{adversarial policies}: an opponent trained explicitly against the victim.
Prior defenses using adversarial training were able to make the victim robust to a specific adversary, but the victim remained vulnerable to new ones.
We conjecture this limitation was due to insufficient \emph{diversity} of adversaries seen during training.
We analyze a defense using population based training to pit the victim against a diverse set of opponents.
We evaluate this defense's robustness against new adversaries in two low-dimensional environments.
This defense increases robustness against adversaries, as measured by the number of attacker training timesteps to exploit the victim. 
Furthermore, we show that robustness is correlated with the size of the opponent population.
\end{abstract}

\section{Introduction}

The discovery of \emph{adversarial examples} \citep{szegedyIntriguing2014} showed that capable image classifiers can be fooled by inputs that are easily classified by a human.
Reinforcement learning (RL) policies have been shown to also be vulnerable to adversarial inputs \cite{huangAdversarial2017, kosDelving2017}.
However, attackers often are not able to directly perturb inputs of a victim agent.
\citet{gleave_adversarial_2020} instead model the attacker as controlling an adversarial agent in an environment shared with the victim.
This adversarial agent does not have any special powers, but can \emph{indirectly} influence the victim's observations by taking actions in the world.
\citeauthor{gleave_adversarial_2020} train such \emph{adversarial policies} against a fixed victim policy using RL, and are able to exploit state-of-the-art victim policies in zero-sum, two-agent, simulated robotics environments \cite{bansal_emergent_2018}.

This vulnerability is surprising given that self-play has produced policies that can defeat the world champions in Go \cite{silverMasteringGameGo2016a} and Dota~2 \cite{openai_dota_2019}.
Critically, zero-sum games are \emph{naturally} adversarial, with self-play being  akin to adversarial training:
so we might expect self-play policies to be naturally robust.
Yet training an adversary for 3\% of the timesteps the original self-play procedure trained for, \citet{gleave_adversarial_2020} manage to exploit the victims.

The resulting adversarial policies exhibit counterintuitive behavior such as falling over and curling into a ball in a game where the goal is to prevent the opponent from getting by. 
This would be a weak strategy against most agents but is able to destabilize the victim.
Similar to adversarial examples, these policies showcase failure modes that affect a model, but which a human would most likely be unaffected by.
Such attacks are a critical danger to deep RL policies in high-stakes settings where there may be adversaries, such as autonomous driving or automated financial trading.

Our work is inspired by prior work in multi-agent RL using populations of agents, such as policy-space response oracles~\citep{lanctot_unified_2017} and population based reinforcement learning~\citep[PBRL;][]{jaderberg_human-level_2019}.
We evaluate using PBRL to train an agent against a diverse population of opponents.
PBRL is a variation of population based training \citep[PBT;][]{jaderberg2017population}, which jointly optimizes a population of models and their hyperparameters for improved convergence, adapted for RL.

Whereas self-play trains an agent to be robust against \emph{itself}, PBRL with a sufficient number of opponents will force an agent to be robust against a wide range of strategies.
We conjecture that this will have a similar benefit as adversarial training \cite{goodfellow2015} for classification models.

We evaluate PBRL as a defense in two simple two-player, zero-sum games.
We find that the self-play baseline can on average be exploited by an adversary using less than 60\% as many timesteps as self-play.
The PBRL-trained policy is more robust: more timesteps are needed until an initial adversarial policy can be found.

We make three key contributions. 
First, we use PBRL as a relatively simple population based method for end-to-end %
robust training in deep RL. 
Second, we evaluate this empirically, finding the PBRL-trained victim is less exploitable. 
Third, we investigate how attributes of the environment (such as dimensionality) and algorithm (such as population size) influence robustness.

\section{Related Work}

\citet{gleave_adversarial_2020} introduced the adversarial policy threat model, and the first attack: using RL to train an adversary against a fixed victim.
\citet{wu_adversarial_2021} added an auxiliary term to reward the victim for paying attention to the adversary.
They apply this attack to the original environments \citep{bansal_emergent_2018}, and a new environment: Roboschool Pong.
\citet{guo_adversarial_2021} develop a different attack for semi-competitive games, exploiting agents in the original environments and Starcraft 2.

Comparatively little attention has been given to defenses relative to attacks.
\citet{gleave_adversarial_2020} attempted to harden the victim by fine-tuning it against a fully trained adversary.
The hardened victim was robust to that particular adversarial policy -- but it was still vulnerable to repeating the attack.
Furthermore, the hardened victim achieved lower performance against the original, non-adversarial opponent.
\citet{guo_adversarial_2021} use the same approach and find that fine-tuning on adversaries from a stronger attack can provide robustness against adversaries from weaker attacks.
However, the authors don't evaluate the robustness of the hardened victim to an attack targeting itself rather than the original victim.

One prior defense is \emph{Adversarially Robust Control} \citep[ARC;][]{kuutti2021arc}.
They consider the semi-competitive setting of autonomous driving.
They find that imitation-learned policies are vulnerable to adversarial vehicles trained to cause collisions -- even when the adversary is limited to only causing preventable collisions.
To improve robustness they fine-tune the imitation policies against an ensemble of adversaries that train concurrently with the main policy.
Since autonomous driving is semi-competitive, an optimal policy against adversaries might fare poorly against regular agents, so \citeauthor{kuutti2021arc} add an auxiliary loss to keep the fine-tuned policy similar to the imitation policy.
In contrast, we focus on the more challenging zero-sum setting which self-play was designed to work with.

\citet{vinitsky2020robust} formulate the single-agent robust RL problem as a zero-sum two-agent problem:
The adversary learns to apply noise perturbations to the environment dynamics trying to minimize the performance of the robust agent in the environment.
To improve robustness the authors train the control agent against a population of these adversaries and find this increases robustness against overfitting to particular adversaries, increases robustness against an out-of-distribution held-out test set, and is less brittle than the domain randomization baseline.

Compared to this, our work considers a more general attack setting:
Instead of an adversarial formulation of a single-agent control setting, we analyze two-agent symmetric and asymmetric games.
It seems likely that these kinds of two-player games have a higher diversity of possible adversarial strategies compared to adversaries that control limited perturbations of transition dynamics.
Lower strategic diversity could be one possible reason why \citeauthor{vinitsky2020robust} encounter diminishing returns at fairly low sizes of adversarial populations.
Additionally, we generally train with higher population sizes, at the cost of increased computational effort.
Furthermore, \citeauthor{vinitsky2020robust} investigate robustness under swaps of adversaries from different training runs, essentially off-distribution robustness, whereas our threat model consists of newly-trained adversaries.

Our work is inspired by prior work in multi-agent RL using populations of agents.
Notably, policy-space response oracles~\citep[PSRO;][]{lanctot_unified_2017} learn an approximate best response to a mixture of policies.
Furthermore, AlphaStar's Nash league~\citep{vinyals_2019} uses a large population of agents playing against each other, with diverse objectives.
Our contribution relative to prior work is an empirical study of the exploitability of population-based trained agents relative to self-play.
To avoid confounders, we keep training as close as possible to our self-play baseline, using a relatively simple form of population-based training compared to much prior work.

There are also a variety of multi-agent RL approaches that do not use populations but may produce policies that are less exploitable than self-play.
One successful technique is counterfactual regret minimization~\citep{zinkevich2007} that can beat professional human poker players~\citep{brown2018}, although has difficulty scaling to high-dimensional state spaces.
We choose to focus on self-play as our baseline since it is widely used and, despite theoretical deficiencies, has produced state-of-the-art results in many environments.

Self-play has been shown to converge to Nash equilibria provided that RL converges to a best response asymptotically \citep{heinrich_selfplay_2015}.
However, RL may never converge to a best response.
A major culprit is non-transitivity: \citet{balduzzi_openended_2019} show that in games like rock-paper-scissors, self-play may get stuck in a cycle.
Even in a transitive game, deep RL algorithms may never converge to a best response if they are stuck in a local minimum, or if the policy network cannot represent the optimal policy.

\section{Background}
Following \citet{gleave_adversarial_2020}, we model the tasks as Markov games $M = (\mathcal{S}, (\mathcal{A}_\alpha, \mathcal{A}_\nu), \mu, \mathcal{T}, \gamma, (R_\alpha, R_\nu))$ \citep{Littman94markovgames} with state set $\mathcal{S}$, action sets $\mathcal{A}_i$, initial state distribution $\mu$, transition dynamics $\mathcal{T}$, discount factor $\gamma$ and reward functions $R_i$.
The index $i \in \{\alpha,\nu\}$ represents the adversarial ($\alpha$) and victim ($\nu$) agent respectively.

We focus on two-player, zero-sum games as they have a clear competitive setup and obvious adversarial goals.
Consequently $R_{\alpha} = -R_{\nu}$.
Notably, the victim policy trained in a zero-sum game will be \emph{harder} to exploit than those in a positive-sum setting, where they may have learned to cooperate with other agents (in an exploitable fashion) to maximize their overall reward.

The attacker has grey-box access to the victim.
That is, they can train against a fixed victim, but cannot directly inspect its weights.
The attacker has no additional capabilities to manipulate the victim or the environment.

\section{PBRL Defense}
We would like to find a Nash equilibrium $(\pi_{\nu}, \pi_{\alpha})$, which for zero-sum games corresponds to the minimax solutions:
We take the $\arg \min_{\pi_\nu}\max_{\pi_\alpha}$ of the expectation $\mathbb{E}\left[\sum_{t=0}^\infty \gamma^tR_{\alpha}(S^{(t)},A_\alpha^{(t)}, A_\nu^{(t)}, S^{(t+1)}) \mid \pi_\alpha,\pi_\nu\right],
$
with random variables sampled from $S^{(0)} \sim \mu$, $A_i^{(t)} \sim \pi_i(\cdot \mid S^{(t)})$, and $S^{(t+1)} \sim \mathcal{T}(S^{(t)}, A_{\alpha}^{(t)}, A_{\nu}^{(t)})$. 

However, as previously discussed self-play may not find the Nash equilibrium.
Yet the self-play policies are nonetheless often highly capable -- against their self-play opponent.
It is known that self-play can get stuck in a \emph{local} Nash equilibrium: population-based training might avoid this failure mode by a greater diversity of opponents.
Notably, adversarially training against some $\pi_{\alpha}$ trained against $\pi_{\nu}$ might just cause $\pi_{\nu}$ to move to a new local Nash equilibrium that is robust to $\pi_{\alpha}$, but not unseen adversary $\pi_{\alpha}'$.

Since population based reinforcement learning \citep[PBRL;][]{jaderberg_human-level_2019} exposes the victim to a broader range of opponents than self-play, it seems likely that it might overcome this limitation while retaining much of self-play's algorithmic simplicity.
We train an agent, the \emph{protagonist}, for robustness by training it against a population of $n$ opponents $\pi_{o_i}$.
By jointly optimizing against multiple opponents we increase the coverage of the space of opponent policies. 
Since an adversary $\pi_\alpha'$ optimizes in a similar way as the opponents it is likely to be close to one of the opponent policies $\pi_{o_i}$.
Moreover, given sufficient diversity in opponents it may be easier for the protagonist to learn a policy close to global Nash than to learn $n$ strategies that overfit to each opponent.

Each of the $n$ opponents has identical architecture and training objective, differing only in the seed used to randomly initialize their network.
We alternate between training the opponents against a fixed protagonist, and a protagonist against all fixed opponents.
Instead of training adversaries such that \emph{in aggregate} they receive as many timesteps as the main agent, all agents (opponent and protagonist) are trained for the same total number of timesteps each.
This alleviates a potential factor for diminishing returns with high population sizes, which \citet{vinitsky2020robust} encountered.
This allows us to make use of an order of magnitude larger population sizes with populations containing tens of opponents.

The total number of training timesteps for all policies is $n+1$ times the number of timesteps the protagonist is trained for. 
When logging timesteps for PBRL training, we report the number of timesteps the protagonist agent trains, since this is the relevant metric for protagonist training.
Note that this means the compute necessary for training PBRL is $n+1$ times higher than self-play at the same number of training steps (although PBRL is more parallelizable).

\def\rllib/{\texttt{RLlib}}

We train all policies -- self-play, PBRL and adversarial -- using Proximal Policy Optimization \citep[PPO;][]{schulman_proximal_2017}. %
PPO is widely used and has achieved good results with self-play in complex environments \citep{bansal_emergent_2018}.
Furthermore, prior work on adversarial policies in higher-dimensional two-agent environments uses PPO \citep{gleave_adversarial_2020}.
We use the PPO implementation in \rllib/~\citep{rllib}, from the \texttt{ray} library \citep{ray}, due to its support for multi-agent environments and parallelizing RL training.

\section{Experiments}
\label{sec:experiments}

We evaluate the PBRL defense in two low-dimensional environments, described in Section~\ref{sec:env}. 
In Section~\ref{sec:attack} we confirm that baseline self-play policies are vulnerable to attacks.
To the best of our knowledge, these are the lowest-dimensional environments in which an adversarial policy has been found.
Finally, in Section~\ref{sec:defense} we find that PBRL improves robustness against adversarial policies, and explore the relationship with population size.

Unless otherwise noted, in all experiments we train 5 seeds of victim policies.
We attack each victim using 3 seeds of adversaries for a total of 15 adversaries.
Unless omitted for legibility, 95\% confidence intervals are shown as shaded regions for training curves and bars in bar plots.

\subsection{Environments}
\label{sec:env}

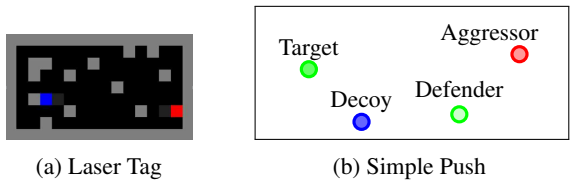
\begin{figure}
    \begin{subfigure}[b]{0.5\linewidth}
        \centering
        \includegraphics[width=0.6\linewidth]{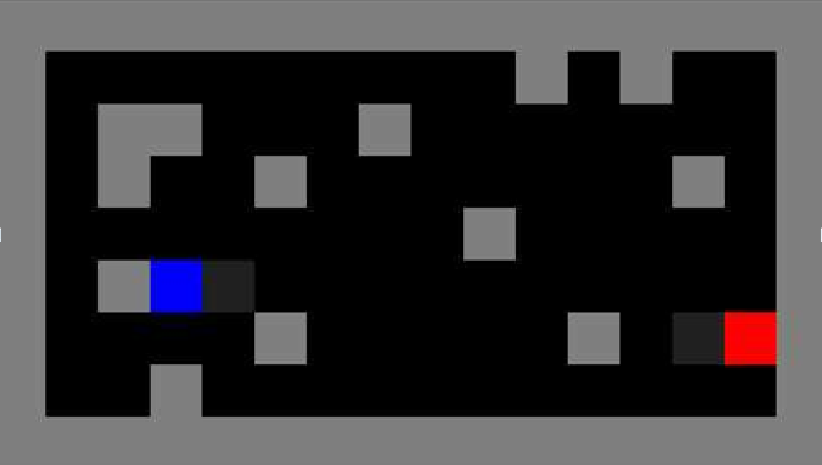}
        \caption{Laser Tag}
        \label{fig:envs:laser}
    \end{subfigure}%
    \begin{subfigure}[b]{0.5\linewidth}
        \centering
        \fbox{
        \begin{tikzpicture}

\node[above ] at (1.7,-0.7) {\footnotesize Aggressor};
\filldraw[color=red!100, fill=red!40, very thick](2.1,-0.7) circle (0.1);
\node[above] at (-0.7,-0.9) {\footnotesize Target};
\filldraw[color=green!100, fill=green!60, very thick] (-0.7,-0.9) circle (0.1);
\node[above] at (1.3,-1.4) {\footnotesize Defender};
\filldraw[color=green!100, fill=green!20, very thick](1.3,-1.5) circle (0.1);
\node[above] at (0,-1.6) {\footnotesize Decoy};
\filldraw[color=blue!100, fill=blue!60, very thick](0,-1.6) circle (0.1);
\end{tikzpicture}
        }
        \caption{Simple Push}
        \label{fig:envs:push}
    \end{subfigure}
    \caption{Illustrations of the (a)~Laser Tag and (b)~Simple Push environments. 
    Laser Tag is a symmetric grid-world game, Simple Push is continuous control and asymmetric.
    It consists of \emph{aggressor} and \emph{defender} agents, as well as target and decoy landmarks. 
    }
    \label{fig:envs}
\end{figure}

We evaluate in two low-dimensional, two-player zero-sum games illustrated in Figure~\ref{fig:envs}: \emph{Laser Tag} and \emph{Simple Push}.

\emph{Laser Tag} is a symmetric game with incomplete information \citep{lanctot_unified_2017}.
The players see 17 spaces in front, 10 to the sides, and 2 spaces behind their agent.
The two agents move on a grid world and get points for tagging each other with a light beam.
Obstacles block movement and beams.
We make the environment zero-sum by also subtracting a point from the tagged player.

\emph{Simple Push} is a continuous environment introduced by \citet{mordatch2017emergence} and released with \citet{lowe2017multi}.
The environment is asymmetric, with one agent the \emph{aggressor} and the other the \emph{defender}.\footnote{Note, that the notion of \emph{aggressor} and \emph{defender} in this environment is orthogonal to \emph{adversary} and \emph{victim} in the sense of adversarial policies.}

The environment contains two randomly placed landmarks. 
Only the defender knows which of these is the true target, the other landmark acts as a \enquote{decoy} for the aggressor.
The aggressor receives positive reward based on the defenders distance to the true target.
Subtracted from this is a relative penalty, based on its own distance.
Unlike vanilla Simple Push, where the defender's rewards are solely based on its own distance, we make the environment zero-sum by giving the defender the negative of the aggressor's reward.

Initial experiments, whose training curves can be found in Figure~\ref{fig:sp_no_comm} of the appendix, 
showed that the attack from \citet{gleave_adversarial_2020} does not find an adversarial policy in vanilla Simple Push.
As Simple Push is very low dimensional (a two-dimensional continuous control task), we develop a variant with a \enquote{cheap talk} communication channel (see Section~\ref{sec:apdx-comm} in the appendix) that increases the dimensionality but does not otherwise change the dynamics.
This channel extends the action and observation spaces allowing agents to send one-hot coded tokens to each other.
With this channel we successfully find adversarial policies, supporting the hypothesis mentioned by \citeauthor{gleave_adversarial_2020}, that increased dimensionality increases the chance to find adversarial policies.
Consequently, we perform experiments in Simple Push with a one-hot coded communication channel of 50 tokens. 
See Section~\ref{sec:apdx-comm} of the appendix for details on the communication channel.

\begin{figure*}[h!]
    \centering
    \begin{minipage}[]{0.33\linewidth}
        \centering
        \includegraphics[width=0.94\columnwidth]{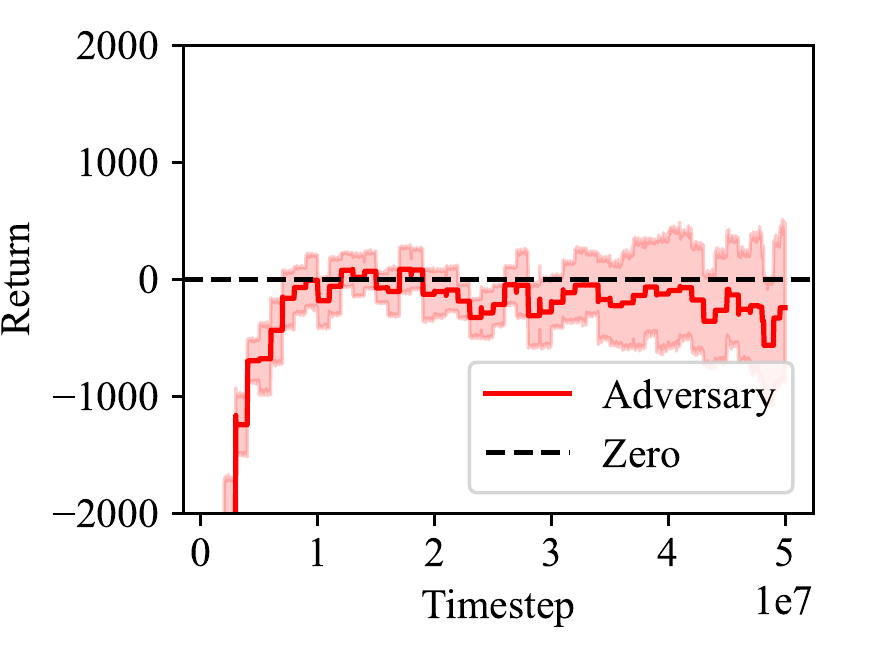}
        \caption{Training curve of adversaries in \emph{Laser Tag}. After fewer than 15 million timesteps of training on average a successful adversarial policy can be found. However variance is high and adversary performance deteriorates against some strong victims over time.}
        \label{fig:lt_attack_sp}
    \end{minipage}%
    \quad%
    \begin{minipage}[]{0.64\linewidth}
        \begin{subfigure}[t]{0.5\linewidth}
            \centering
            \includegraphics[width=\linewidth]{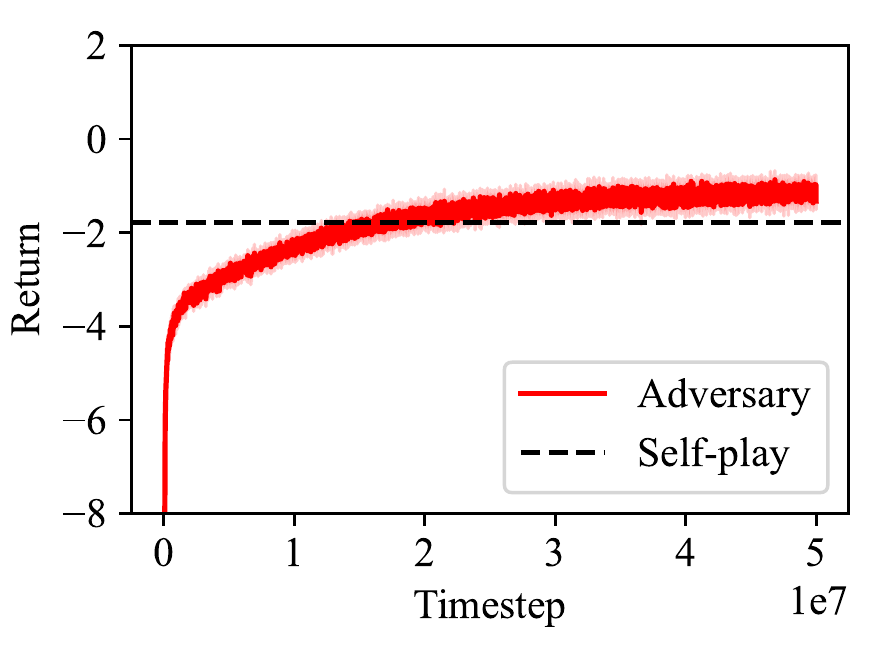}
            \caption{Aggressor return.}
            \label{fig:sp_agressor}
        \end{subfigure}%
        \begin{subfigure}[t]{0.5\linewidth}
            \centering
            \includegraphics[width=\linewidth]{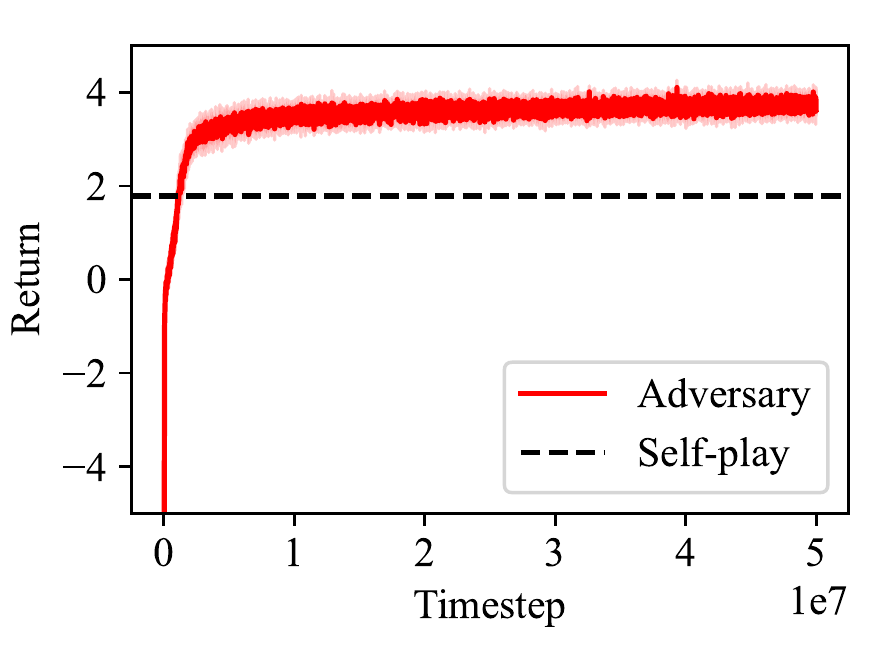}
        \caption{Defender return.}
            \label{fig:sp_defender}
        \end{subfigure}
        \vspace{4px}
        \caption{\emph{Simple Push} average return of adversary controlling aggressor (left) and defender (right) \emph{with} 50-token communication channel.
        The black dotted line marks the return achieved by the self-play training baseline controlling the respective agent at the end of training.
        Adversary controlling aggressor leads to weak attack, defender leads to strong attack.}
        \label{fig:sp}
    \end{minipage}
\end{figure*}

\subsection{Self-play Baseline}
\label{sec:attack}

Before evaluating our PBRL defense, we first consider the exploitability of the self-play baseline by the standard adversarial policy attack from \citet{gleave_adversarial_2020}.
We find self-play policies in both Laser Tag and Simple Push to be vulnerable.
However, in Simple Push the attack only succeeds when there is a communication channel.
In Laser Tag adversarial policies can be found, however variance is high and some self-play victims are hard to attack.

These results add nuance to the conclusion from \citet{gleave_adversarial_2020} that adversarial policies are easier to find in higher-dimensional environments.
Our environments are significantly lower-dimensional than those considered by \citeauthor{gleave_adversarial_2020}, suggesting the minimum dimensionality for attack is smaller than previously believed.
However, the fact that policies are only vulnerable in Simple Push given a communication channel, and the high variance in the victim robustness in Laser Tag, supports the overall claim that dimensionality is an important mediator for exploitability.

\paragraph{Laser Tag.}
\label{sec:attack-laser}

We train self-play policies in the symmetric Laser Tag environment for 25 million timesteps.
Figure~\ref{fig:lt_attack_sp} shows the average return of the adversaries trained against these victims.
We train adversaries for 50 million timesteps, twice as many as the victims, in order to reason about the adversaries' behavior given more compute.
Since the game is symmetric, an agent with a return above zero outperforms its opponent.

Successful adversarial policies can be found: if the adversary were to stop training once it outperforms the victim, on average fewer than 15 million timesteps are necessary for a successful attack. %
The loose confidence interval suggests high variability in different seeds.
Some of the trained victims are robust while others are not.
On average attacker performance deteriorates after 20 million timesteps which suggests an instability in adversary training.

\paragraph{Simple Push.}
\label{sec:attack-push}

Since Simple Push has an asymmetric observation space, we train self-play using a separate policy for either player.
We train the agents for 25 million timesteps, which we expect to be more than sufficient given the 625,000 timesteps used in prior work \cite{lowe2017multi}.
While prior work used MADDPG, not PPO, in exploratory experiments we find PPO to perform comparably to MADDPG.%

Again we train adversaries for 50 million timesteps, twice as many as the victims, in order to reason about the adversaries' behavior given more compute.
Figure~\ref{fig:sp_defender} shows when the adversary controls the defender the adversary achieves almost twice the return on average (in red) as the self-play baseline (in black) at 25 million timesteps, which is the point where adversary and victim trained for the same number of timesteps.
By contrast, Figure~\ref{fig:sp_agressor} shows that when controlling the aggressor the adversary needs around 20 million timesteps just to match the victim. 
Return after training the adversary for twice as many timesteps as the victim is only slightly higher than the baseline. 

These results suggest defender policies in Simple Push are more robust to adversarial policies than aggressor policies.
We conjecture this asymmetry is due to the defender having more information than the aggressor: it knows the target landmark.
Consequently, %
the aggressor needs to observe the opponent to learn the true target landmark.
The defender could exploit this and perform movements that fool a victim aggressor.
Due to the stronger nature of the attack controlling the defender, we focus our defense in the upcoming section on adversaries that control the defender agent.

\subsection{PBRL Defense}
\label{sec:defense}

\begin{figure*}[h!]
    \centering
    \begin{subfigure}[t]{0.43\linewidth}
            \includegraphics[width=0.99\linewidth]{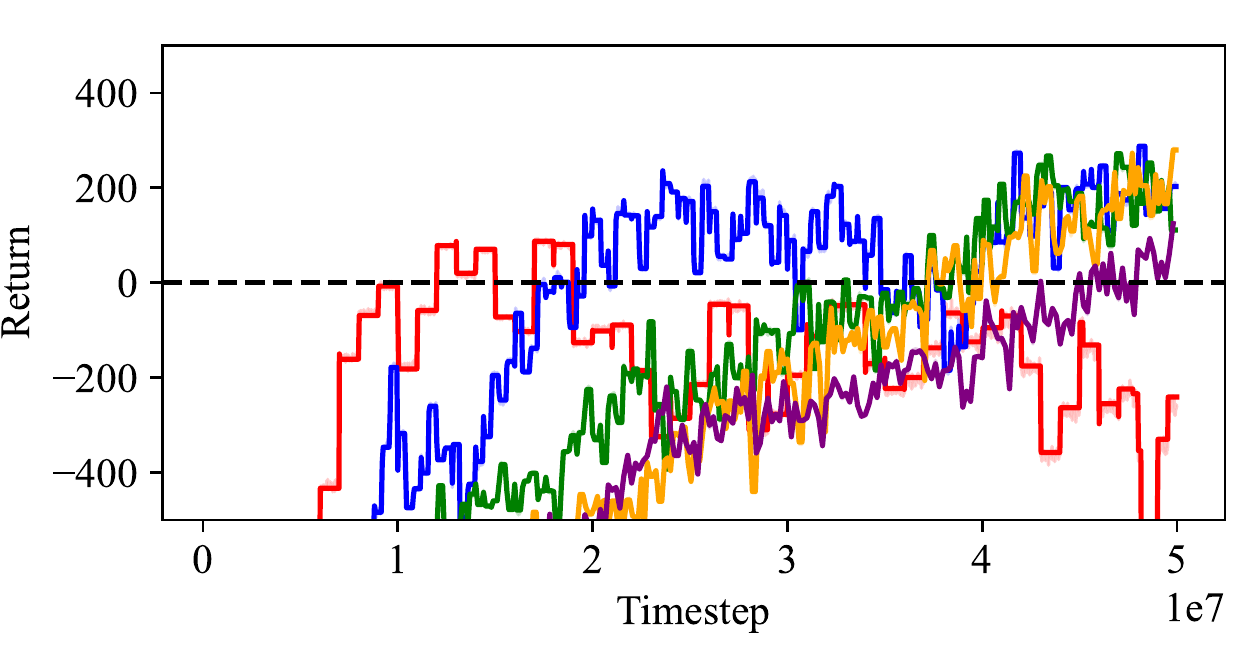}
        \caption{Averages of adversary returns.}
        \label{fig:curve-laser-pbt}
    \end{subfigure}%
    \begin{subfigure}[t]{0.57\linewidth}
        \includegraphics[width=0.254\linewidth]{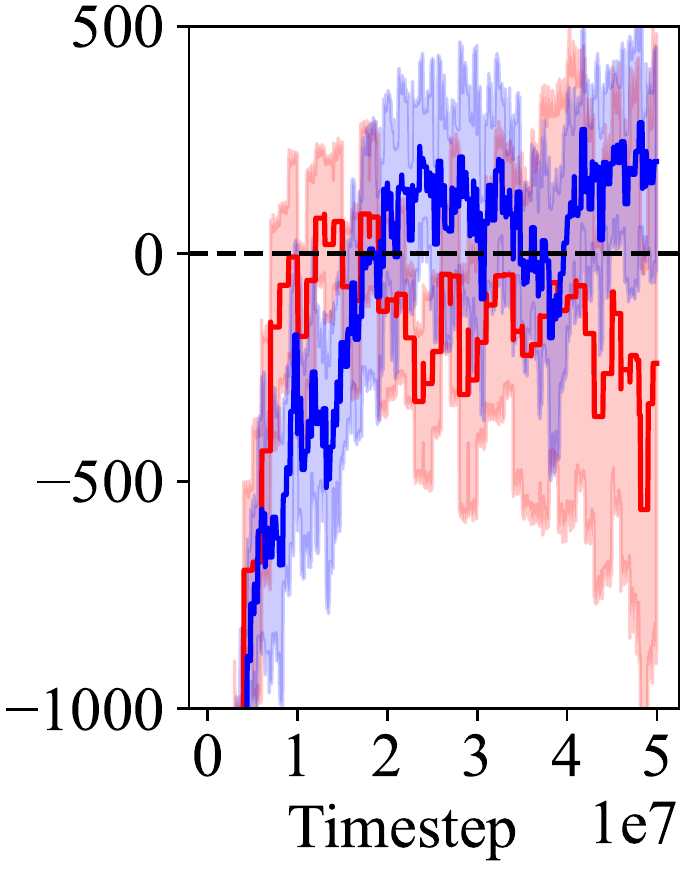}%
        \includegraphics[width=0.199\linewidth]{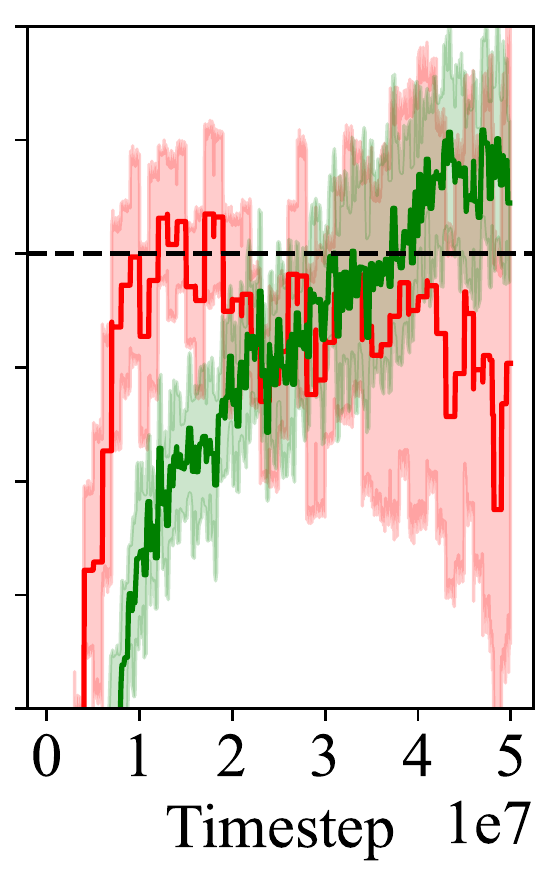}%
        \includegraphics[width=0.199\linewidth]{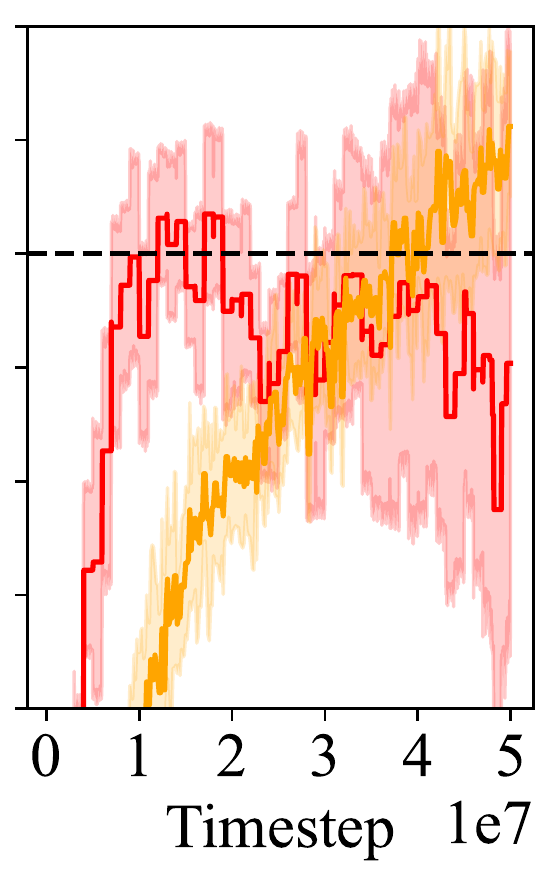}%
        \includegraphics[width=0.199\linewidth]{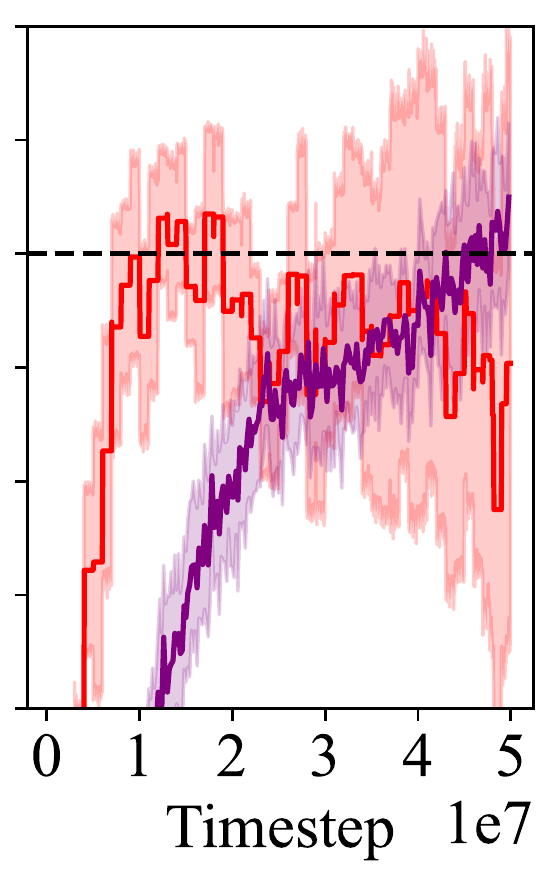}%
        \includegraphics[width=0.15\linewidth]{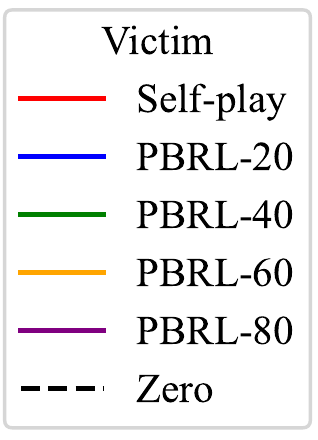}%
    \caption{Averages with confidence intervals.}
    \label{fig:curve-laser-pbt-ci}
\end{subfigure}
\caption{Training curves for adversaries attacking in \emph{Laser Tag}.
We measure exploitability by comparing how many timesteps the adversary needs to train for until it outperforms the victim, i.e. achieves larger than 0 return. We compare the self-play baseline and policies PBRL-$n$, trained with population size $n$.
With increasing population size the number of timesteps the adversary needs increases, suggesting that exploitability decreases with higher population sizes.
However, we also find self-play is more robust than policies trained using PBRL against adversaries that continue training, suggesting that there might be more instability when training an adversary against self-play. Figures in (b) show the confidence intervals for returns of adversaries attacking PBRL and self-play policies from (a).}
\end{figure*}

In this section, we evaluate the effectiveness of a PBRL defense by trying to exploit PBRL-trained policies.
We find some improvement in robustness relative to the self-play baseline in both environments.
In Laser Tag larger populations increase the number of timesteps needed to find the first adversarial policy -- at the cost of requiring more computational resources. 
In Simple Push adversaries don't outperform self-play baselines and there are no significant improvements with more than $n=2$ opponents.

We evaluate the defense by attacking the PBRL-trained protagonist agent, and compare to the attacks on the baseline self-play policies from Section~\ref{sec:attack}.
We focus on the relative performance of the adversary compared to what the victims achieve on-distribution. %
We continue training adversaries up to 50 million timesteps against the same fixed 25-million-timestep victim to evaluate how many timesteps are needed to attack more robust victims.

\paragraph{Laser Tag.}

To explore the impact of population size we train policies with $n=20$, $40$, $60$, and $80$ opponents in Laser Tag.
Figure~\ref{fig:curve-laser-pbt} shows the average return of adversaries attacking these hardened protagonists.
We find that using PBRL increases robustness: finding an adversary that achieves higher than 0 reward takes more timesteps on average.
While fairly noisy, generally the number of timesteps needed to outperform the victim -- when return crosses over the zero line -- increases with increasing population size.
An adversary attacking a protagonist hardened against a population of size 80 needs to train on almost double the timesteps as the self-play victim.
However, the relative difference in timesteps is smaller for higher population sizes.

Although the adversarial policy was trained for up to double the number of timesteps as the protagonist agent, note that PBRL used $80$ times as much compute for every timestep of the protagonist, since it had to train the opponents for the same number of timesteps.
Additionally, we find that adversaries that continue to train eventually do outperform PBRL victims on average, whereas average performance against self-play victims \emph{decreases} over time.
When evaluating $95\%$ confidence intervals (Figure~\ref{fig:curve-laser-pbt-ci}), %
variance of adversaries attacking self-play increases over time, which suggests adversary training to be less stable when attacking self-play as opposed to attacking PBRL.
Self-play seems to converge to policies of widely varying robustness, whereas in PBRL variance of attackers is lower.

\paragraph{Simple Push.}
\label{sec:defense-push}

We focus on making the \emph{aggressor} agent more robust in Simple Push, as Section~\ref{sec:attack-push} showed that the defender self-play policy is already relatively robust to attack.
Consequently, the protagonist controls the \emph{aggressor} and adversaries the \emph{defender} agent.
We use PBRL to train against $n=$~2, 4, 8 and 16 opponents.
Since the environment is not balanced, as a baseline we judge performance by comparing to the returns at the end of training.
The adversary is successful if it achieves higher results against the victim than the PBRL opponents did on average at the end of training against that same victim policy.

\begin{figure}[h]
    \centering
    \includegraphics[width=0.85\columnwidth]{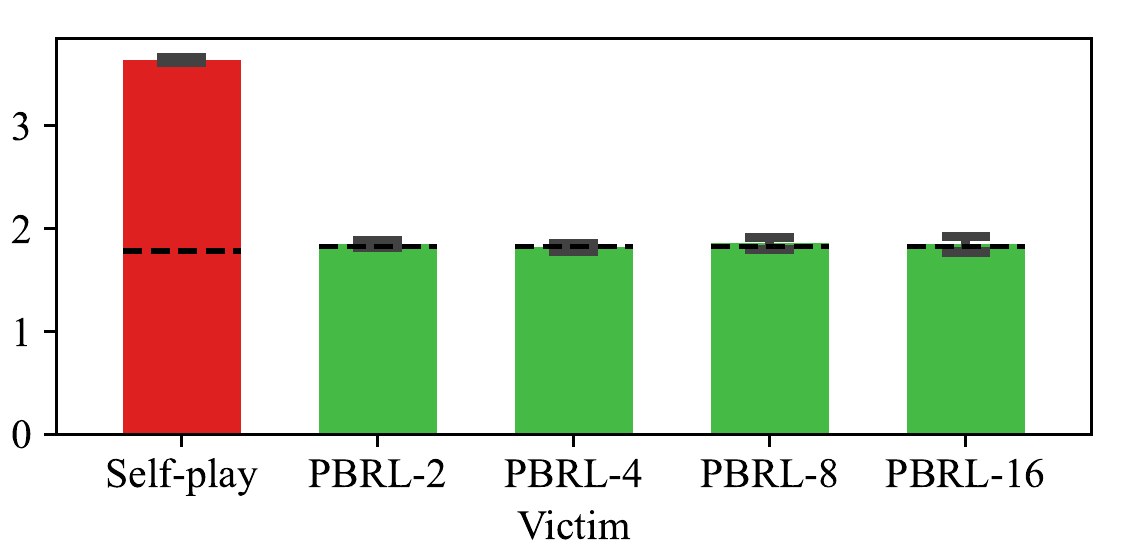}
\caption{PBRL-trained agents (green) are significantly more robust to adversarial attack than the self-play baseline (red), and attain similar return to the baseline that plays against a \emph{non-adversarial} agent (black dotted line). 
}
     \label{fig:push_pbt}
\end{figure}

Figure~\ref{fig:push_pbt} shows the returns at 25 million timesteps.
Since baselines in different setups could converge to different returns, we calculate separate baseline thresholds for each of the 4 settings (in addition to the self-play baseline from Figure~\ref{fig:sp}), marked by the dashed black line.
Notably, the values all policies converge to differ by less than 3\%.
The PBRL-trained agent is significantly more robust than the self-play policy, in red.
In fact, the PBRL protagonist achieves similar return \emph{under a zero-shot attack} as the self-play policy does \emph{against its self-play opponent} (the dashed threshold).

\begin{figure}[h]
    \centering
     \includegraphics[width=0.9\columnwidth]{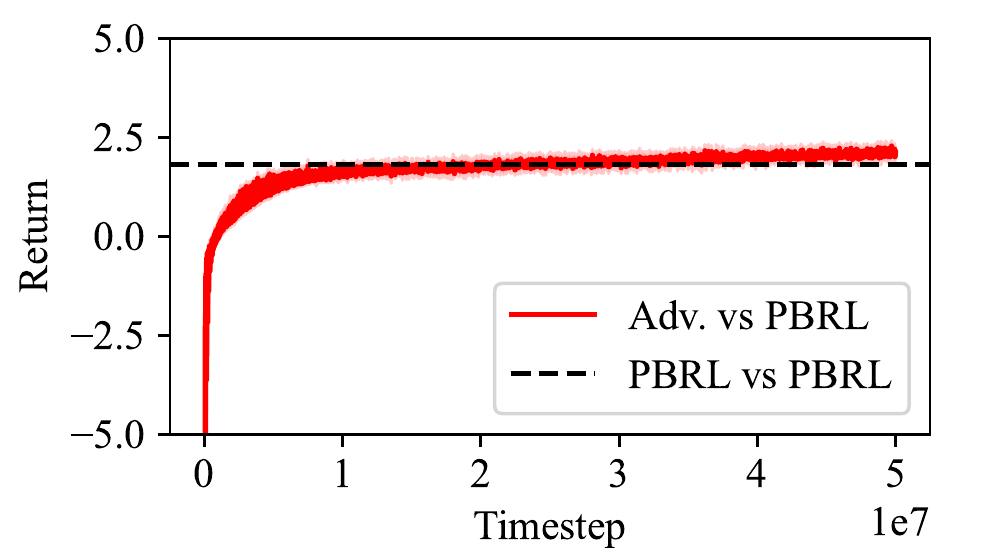}
    \caption{Return of adversaries attacking all PBRL-trained agents (number of opponents $n = 2, 4,8,16$), averaged over all 60 adversaries (15 per $n$). Victims are all similarly robust in Simple Push, adversaries need at least as many timesteps as the victim in order to outperform it.}
     \label{fig:push_pbt_curve}
\end{figure}

Figure~\ref{fig:push_pbt_curve} shows the training curve, when training adversaries for up to 50 million timesteps.
Since there is no discernible difference in the 4 PBRL settings, we average over these for a total of 60 adversarial policies.
When the adversary trains \emph{twice as much} as the PBRL protagonist victim, the attack outperforms the victim, but is not particularly strong.

Although PBRL is significantly more robust than self-play (effectively PBRL with $n=1$ opponents), perhaps surprisingly there is little benefit from using more than $n=2$ opponents.
In particular, there is no clear decrease in robustness when using $n$ as low as 2, which is the lowest PBRL settings that does not degenerate to self-play.

This is in contrast to Laser Tag, which saw large differences in robustness depending on population size.
This is likely due to different environmental dynamics.
In Simple Push there are only a handful of high-level strategies that one can pursue.
By contrast, the Laser Tag environment allows for more variation in the details of possible strategies, making it harder to achieve full robustness.
Additionally, it is possible that the intervention of slightly increasing dimensionality, by adding a \enquote{cheap-talk} channel, can be circumvented with minimally higher diversity during training.

\section{Discussion and Future Work}

In this work, we evaluated a hardened PBRL victim agent against adversarial policies trained with similar numbers of timesteps of experience as the victim agent.
However, PBRL \emph{also} requires training $n$ opponents for this many timesteps -- so the \emph{total} number of timesteps and computational resources is $n+1$ times greater.

We believe this overhead is often tolerable.
First, defenders may be able to limit the number of timesteps an attacker can train against the victim, such as if access to the policy is behind a rate-limited API.
Second, the number of opponents $n$ can sometimes be quite small -- just $n=2$ suffices for Simple Push.
Finally, defenders often have significant computational resources: while PBRL is unlikely to prevent an attack from a sophisticated adversary like a nation-state, it may be enough to defeat many low-resource attacks.
Nonetheless, reducing this computational overhead is an important direction for future work.
For example, can we obtain similar performance with fewer opponents if we train them to be maximally diverse from one another?

In addition, \emph{if} additional compute resources are available, this approach allows a defender to \emph{make use} of them.
Once an agent has converged, using additional compute to continue training in self-play is usually of no use.
However, convergence is not sufficient for robustness -- as illustrated by the existence of adversarial policies.
This approach allows a purposeful use of additional computing power in a way that is fairly similar to self-play.
However, more advanced algorithmic improvements such as PSRO~\citep{lanctot_unified_2017} might be able to provide similar benefits with a smaller increase in compute.

A key open question is how the number of opponents $n$ required for robustness scales with the complexity of the environment.
PBRL will scale poorly if the required population size is proportional to the size of the state space: in more complex environments each opponent will take longer to train \emph{and} more opponents will be required.
But a priori it seems likely that $n$ may depend more on the number of high-level strategies in the environment.
This is only loosely related to the dimensionality of the state space.
For example, some simple matrix games have high strategic complexity, while some high-dimensional video games have only a handful of sensible strategies.

Our evaluation uses the original adversarial policy attack of \citet{gleave_adversarial_2020}, which we established was strong enough to exploit unhardened victims in these environments.
However, it is possible that alternative attacks would be able to exploit even our hardened victim.
We hope to see iterative development of stronger attacks and defenses, similar to the trend in adversarial examples more broadly.

\section{Conclusion}

We evaluate a population-based defense to reduce the exploitability of RL policies.
Empirically, we see an increase in zero-shot robustness against new adversaries compared to self-play training.
However, some self-play victims are naturally robust and PBRL comes with an increased computational cost.
We find the required size of the population depends on the environment used, and larger populations can increase overall robustness.
Supplementary material available at \hyperref[]{\url{https://reducing-exploitability.github.io}}, source code available at \hyperref[]{\url{https://github.com/HumanCompatibleAI/reducing-exploitability}}.

\section*{Acknowledgements}
\ifdefined\isaccepted
We thank Philip Sperl and researchers at the Fraunhofer Institute for Applied and Integrated Security and at the Center for Human-Compatible AI for helpful discussions.
We thank P.~A.~M. Casares for helpful feedback on earlier drafts of this paper.
We also thank our anonymous reviewers for feedback on this paper.
Work supported in part by the Berkeley Existential Risk Initative and Centre for Effective Altruism Long Term Future Fund.
\else
Removed for double blind review.
\fi

\bibliography{ref}
\bibliographystyle{icml2022}

\newpage
\appendix
\onecolumn

\section{Adding a Communication Channel to Simple Push}
\label{sec:apdx-comm}

\begin{figure}[h]
    \begin{subfigure}[b]{0.5\linewidth}
        \centering
        \includegraphics[width=0.8\linewidth]{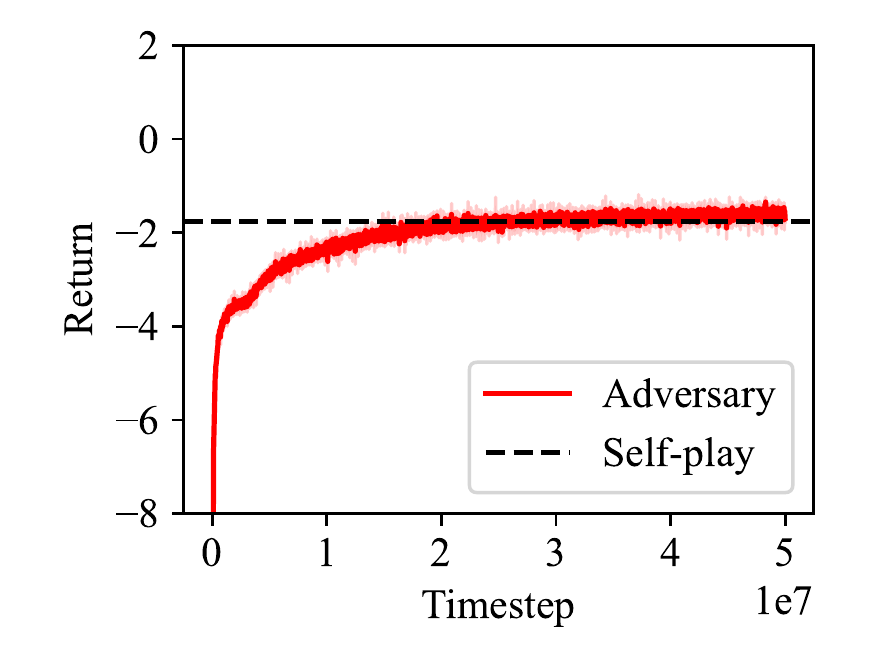}
	\caption{Agressor.}
    \label{fig:no_comm_0}
    \end{subfigure}%
    \begin{subfigure}[b]{0.5\linewidth}
        \centering
        \includegraphics[width=0.8\linewidth]{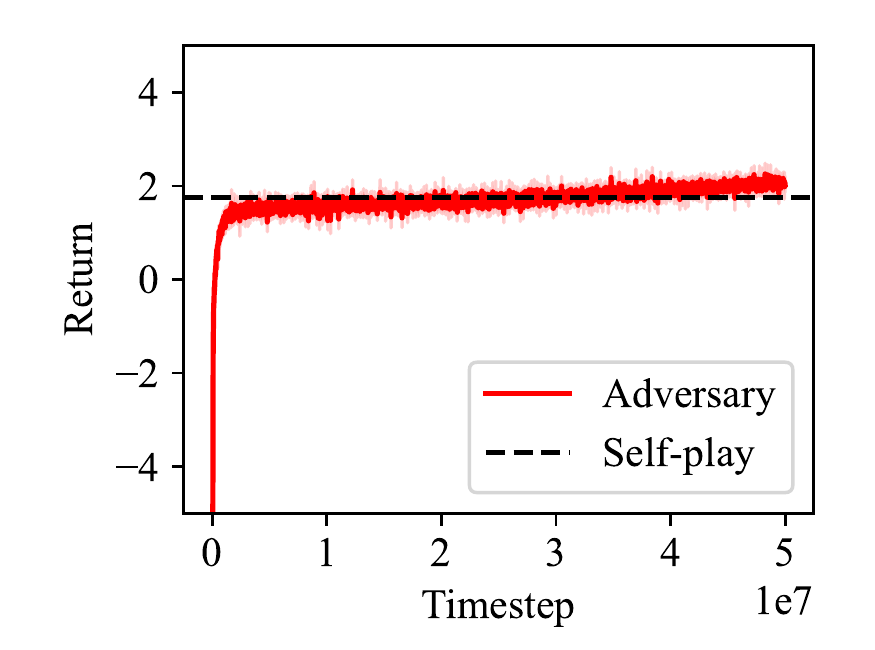}
	\caption{Defender.}
    \label{fig:no_comm_1}
    \end{subfigure}
    \caption{Average return of agent controlling (a) agressor and (b) defender in Simple Push \emph{without} communication. The adversary, in red, fails to outperform the self-play baseline, in black.}%
    \label{fig:sp_no_comm}
\end{figure}

Our initial experiments in the \emph{Simple Push} environment did not lead to adversarial policies that were capable of outperforming their victims (See Figures~\ref{fig:no_comm_0}, \ref{fig:no_comm_1}).

Inspired by the cooperative environments explained in \citet{lowe2017multi}, we add a communication channel: this channel allows each agent to observe a one-hot coded action taken by the other agent.
This communication channel has no other effect on environment dynamics, and agents' reward does not depend on the contents of the communication channel.
The size of the communication channel essentially represents the number of tokens either agent can use to communicate with the other.
Because this setting is competitive, there is no reason for an agent to provide information in the communication channel which would be beneficial to the opponent.
Therefore, an optimal policy should simply ignore the \enquote{messages} sent by the opponent.
However, this channel increases the dimensionality and offers an adversary the possibility to learn messages that might \enquote{confuse} a (sub-optimal) victim.

In a small ablation on communication channels supporting 10, 25, 50 and 100 tokens, we find that adversarial policies are successful with 50 or more tokens and unsuccessful at less than 25.
We also find that the number of timesteps until convergence, as well as general instability during training increases with higher sizes.
For further experiments we use a communication channel of 50 tokens to allow for fast training while still providing an environment in which adversarial policies are possible.

\end{document}